\documentclass{article}

\usepackage{PRIMEarxiv}

\usepackage[utf8]{inputenc} 
\usepackage[T1]{fontenc}    
\usepackage{hyperref}       
\usepackage{url}            
\usepackage{booktabs}       
\usepackage{amsfonts}       
\usepackage{nicefrac}       
\usepackage{microtype}      
\usepackage{lipsum}
\usepackage{fancyhdr}       
\usepackage{graphicx,subfigure}      
\graphicspath{{media/}}     

\usepackage{amsmath}
\usepackage{amssymb}
\usepackage{mathtools}
\usepackage{amsthm}
\usepackage{algorithm}
\usepackage{algorithmic}
\usepackage{tikz,graphics,color,fullpage,float,epsf,caption,subcaption}

\theoremstyle{plain}

\theoremstyle{definition}

\theoremstyle{remark}

\usepackage[textsize=tiny]{todonotes}
\usepackage[capitalize,noabbrev]{cleveref}
\usepackage{listings}
\usepackage{xcolor}
\definecolor{dkgreen}{rgb}{0,0.6,0}  
\definecolor{mauve}{rgb}{0.58,0,0.82}  
\title{Learning How To Ask: Cycle-Consistency Refines Prompts in Multimodal Foundation Models
}

\author{
  \textsuperscript{*}Maurice Diesendruck, \textsuperscript{*}Jianzhe Lin, Shima Imani, Gayathri Mahalingam, Mingyang Xu, Jie Zhao \\
  Microsoft Research \\
  \texttt{\{mdiesendruck, jianzhelin, shimaimani, gmahalingam, mingyangxu, zhaojie\}@microsoft.com} \\
}

\begin{document}
\footnote{
* Co-first authors. Alphabetical by last name. \\
Work in progress. Corresponding authors: mdiesendruck@microsoft.com, jianzhelin@microsoft.com}
\maketitle

\begin{abstract}
When LLMs perform zero-shot inference, they typically use
a prompt with a task specification, and generate a completion. However, there is no work to explore the possibility of the reverse -- going from completion to task specification. In this paper, we employ both directions 
to perform \textit{cycle-supervised} learning entirely in-context. 
Our goal is to create a forward map $f: X \rightarrow Y$ (e.g. image $\rightarrow$ generated caption), coupled with a backward map $g: Y \rightarrow X$ (e.g. caption $\rightarrow$ generated image) to construct a cycle-consistency ``loss" (formulated as an update to the prompt) to enforce $g(f(X)) \approx X$. The technique, called \textit{CyclePrompt}, uses cycle-consistency as a free supervisory signal to iteratively craft the prompt. Importantly, CyclePrompt reinforces model performance without expensive fine-tuning, without training data, and 
 without the complexity of external environments (e.g. compilers, APIs).
We demonstrate CyclePrompt in two domains: code generation and image captioning. Our results on the HumanEval coding benchmark put us in first place on the leaderboard among models that do not rely on extra training data or usage of external environments, and third overall. Compared to the GPT4 baseline, we improve accuracy from 80.5\% to 87.2\%. In the vision-language space, we generate detailed image captions which outperform baseline zero-shot GPT4V captions, when tested against natural (VQAv2) and diagrammatic (FigureQA) visual question-answering benchmarks.
To the best of our knowledge, this is the first use of self-supervised learning for prompting. 


\end{abstract}

\begin{figure}[ht]
    \centering
    \includegraphics[width=0.48\textwidth]{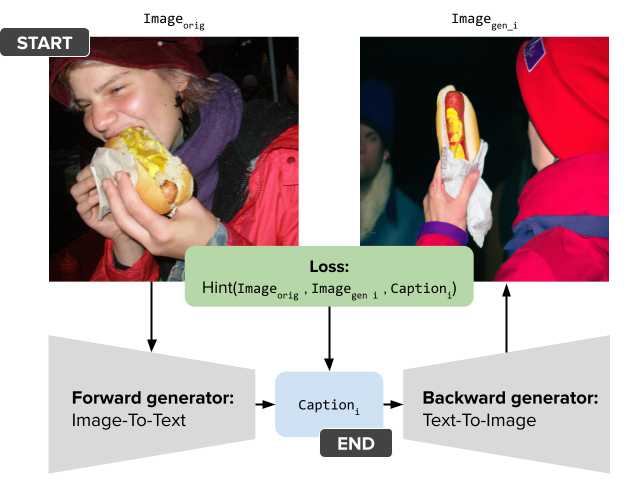}
    \caption{The AutoEncoder-like structure for CyclePrompt. In the example, the \underline{first cycle caption is:} \textit{``A person in a red jacket with a denim sleeve is holding a hot dog with yellow mustard in a bun. They are wearing a purple scarf and a red hat with a flower on it. There is a glimpse of another person in a red jacket in the background. It appears to be nighttime, and there's a white canopy in the background, possibly indicating an outdoor event."}, and \underline{the generated hint is:} \textit{``The person's stance should be slightly bent forward, and the hot dog should be held in both hands with a napkin wrapped around it"}. The hint refines the prompt to improve the generation of the following cycle.}
    \label{fig:structure}
\end{figure}

\section{Introduction}
\label{introduction}
Prompting is the main way people interact with large language models (LLMs). With prompting, people can accomplish various tasks, e.g. code generation, question answering, and context summarization. Typically, the prompt consists of two parts, a task specification and the data to which the task applies; and given how costly it can be to fine-tune LLMs, modifying the prompt is often the best option for controlling model performance.
One method, called ``Reflexion" \cite{shinn2023reflexion}, lets LLMs refine a prompt through verbal reinforcement, based on learning from prior failings. Specifically, Reflexion converts feedback from an external environment (e.g. compilers, APIs) into verbal feedback in the form of a text summary, which is added as additional context to the task specification of the prompt. In doing this, the updated prompt yields better results.

However, a limitation of Reflexion is the availability of an external environment, which can be seen as providing extra training data. To achieve verbal reinforcement \textit{in the absence of external environments}, we propose a free self-supervision based on the concept of cycle-consistency.


This approach is inspired by cycle-consistency in translation tasks, where translating a sentence into another language and back is expected to yield the original sentence. Of course, cycle-consistency is well-known in machine learning, and has been a component of loss functions in image generation \cite{zhu2017unpaired} and in object tracking \cite{wang2019learning}, among others. We for the first time extend this concept to prompting for LLMs, and demonstrate its effectiveness.

\begin{figure*}
    \centering
    \includegraphics[width=0.95\textwidth]{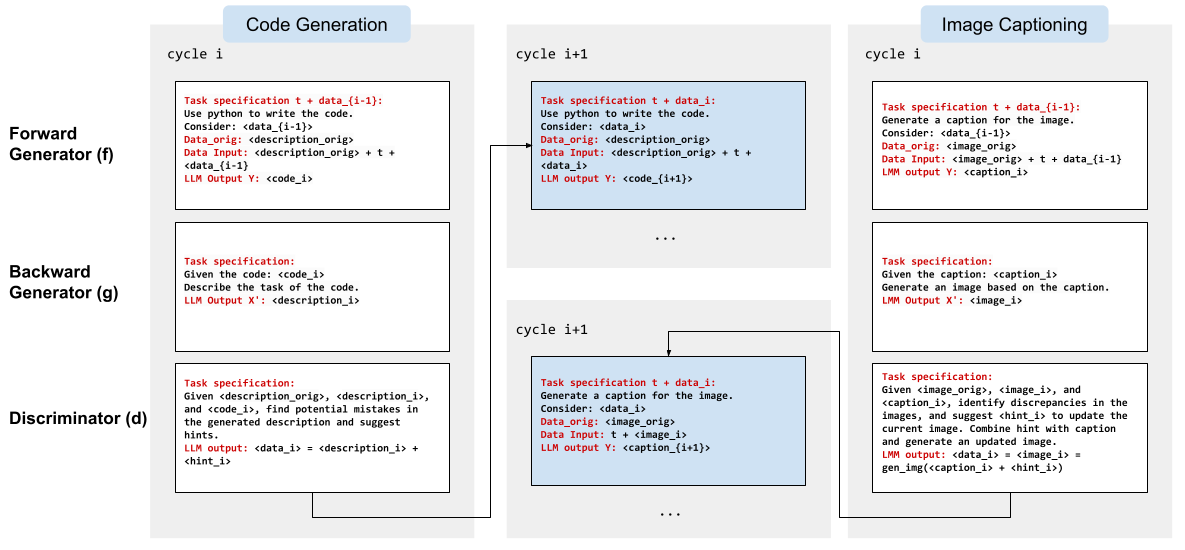}
    \caption{A general flowchart for CyclePrompt, with applications for code generation and image captioning.}
    \label{fig:cycleprompt}
\end{figure*}

To make the cycle-consistency possible, we propose the ``Specification-Completion-Specification" cycle.
Two examples are the Text-Code-Text and Image-Text-Image cycles, which correspond to the code generation and image captioning tasks, respectively. 
Formally, a cycle consists of a forward function $f: X \rightarrow Y$ (e.g. text $\rightarrow$ generated code), and a backward function $g: Y \rightarrow X$ (e.g. code $\rightarrow$ generated text). As can be expected, most pairings of $f$ and $g$ are imperfect, so to improve the cycle, we formulate a cycle-consistency ``loss". This loss compares $x$ and $g(f(X))$ and refines the prompt with the aim of reducing inconsistency. 
This routine is inspired by the autoencoder, though instead of generating a discriminative feature, it uses the cycle as a free supervisory signal to refine the prompt. We call the technique \textit{CyclePrompt}, and visualize its flow in Fig.~\ref{fig:structure} and application to code generation and image captioning in Fig.~\ref{fig:cycleprompt}.
\begin{table}
    \label{table:adv-over-sota}
    \begin{center}
            \begin{tabular}{lcccc}
              Approach & Reasoning &  Self-reflection & External environments & MultiModality \\
              \hline
               CoT \cite{wang2022self}  & \textcolor{green}{$\checkmark$} & \textcolor{red}{$\times$} & \textcolor{green}{$\times$} & \textcolor{red}{$\times$} \\ 
              \hline
               ReAct \cite{yao2022react} & \textcolor{green}{$\checkmark$} & \textcolor{red}{$\times$} & \textcolor{green}{$\times$} & \textcolor{red}{$\times$} \\
              \hline
               ToT \cite{yao2023tree} & \textcolor{green}{$\checkmark$} & \textcolor{red}{$\times$} & \textcolor{green}{$\times$} & \textcolor{red}{$\times$} \\
              \hline
               RAP \cite{hao2023reasoning} & \textcolor{green}{$\checkmark$} & \textcolor{red}{$\times$} & \textcolor{green}{$\times$} & \textcolor{red}{$\times$} \\
               \hline
               Self-Refine \cite{madaan2023self}  & \textcolor{green}{$\checkmark$} & \textcolor{red}{$\times$} & \textcolor{green}{$\times$} & \textcolor{red}{$\times$}\\
              \hline
               Beam Search \cite{pryzant2023automatic} & \textcolor{green}{$\checkmark$} & \textcolor{red}{$\times$} & \textcolor{green}{$\times$} & \textcolor{red}{$\times$} \\
              \hline
               Reflexion \cite{shinn2023reflexion} & \textcolor{green}{$\checkmark$} & \textcolor{green}{$\checkmark$} & \textcolor{red}{$\checkmark$} & \textcolor{red}{$\times$} \\
              \hline
              LATS \cite{zhou2023language} & \textcolor{green}{$\checkmark$} & \textcolor{green}{$\checkmark$} & \textcolor{red}{$\checkmark$} & \textcolor{red}{$\times$} \\
              \hline
               \textbf{CyclePrompt(Ours)} & \textcolor{green}{$\checkmark$} & \textcolor{green}{$\checkmark$} & \textcolor{green}{$\times$} & \textcolor{green}{$\checkmark$} \\
               \hline
            \end{tabular}
    \end{center}
    \caption{A general comparison among state of the arts. CyclePrompt is the first work to achieve self-reflection without the support of external environments. Also, CyclePrompt achieve the reflection across different modalities.}
\end{table}  




In our experiments, when applied to code generation, CyclePrompt achieves third place on the leaderboard of the HumanEval benchmark \cite{chen2021evaluating}, 
and achieves first place among unassisted models. When applied in the vision-language space, CyclePrompt 
produces highly detailed image captions that outperform baseline zero-shot GPT4V captions. These results underscore the potential of in-context, cycle-based reflection and refinement as a simple and powerful tool for optimizing model use. 

Our main contributions are as follows. We:
\begin{itemize}
    \item Introduce the ``Specification $\rightarrow$ Completion $\rightarrow$ Specification" cycle for prompt refinement.
    \item Introduce a ``semantic" cycle-consistency loss that extracts higher model performance without training data, expertise, or external systems.
    \item Demonstrate our approach in code generation, achieving state-of-the-art results on the HumanEval dataset; and in image captioning, where question-answering ability of our captions exceeds that of zero-shot GPT4V captions.
    \item Provide insights into when and why our approach succeeds, including discussion on sensitivity and precision of cycle functions, and discussion about failure modes that present intriguing future directions for characterizing model misalignment.

\end{itemize}

\section{CyclePrompt} 
   
Our approach involves three major components: a forward generator, a backward generator, and a discriminator. These components work together in an iterative refinement process to improve model performance.  
   
The forward generator, denoted as $f$, is responsible for generating an output $y$ given an input data $x$. Note that an input is made of two parts, the task specification $t$ and the input data $s$, yielding $f(x) = y$ where $x = t + s$. In the code generation task, the input data is a coding task and the output is a piece of code; in the image captioning task, the input data is an image and the output is a text caption. The backward generator, denoted as $g$, translates the output of the forward generator back into the original space as $s'$.  
   
The discriminator plays a crucial role in the refinement process, and works iteratively from cycle to cycle. This function takes as input the original data $s$, mapped value $y$, and output from backward generator $s' = g(y) = g(f(t + s))$, and produces a hint, $hint = d(s, s', y)$. The $hint$ is used to update the current value of $s$ for the next round. 
The entire cycle procedure is visualized in Fig. \ref{fig:cycleprompt}. Note that the hint is always generated based on the inconsistency between the currently generated data $s'$ and the original data $s$.

This process is repeated for $N$ cycles to achieve cycle-consistency, i.e. $\texttt{Consistent}(s, s') == \texttt{True}$. The goal is to minimize inconsistency between the original data $s$ and the translated output $s'$, thereby improving the performance of the model. The ultimate goal is to get final output $y_N$. 
   
While a convergence measure could be used to determine when to stop the iterative process, we leave the exploration of such a measure for future work. In this work, we focus on demonstrating the effectiveness of the CyclePrompt approach through a fixed number of cycles (we empirically use 4 cycles for both code generation and image captioning).
   
\subsection{Mathematical Formulation}
The following mathematical formulation more closely specifies the function inputs and outputs, noting the cycle indices. We define forward generator $f: X \rightarrow Y$, backward generator $g: Y \rightarrow X$, and discriminator $d: X \times X \times Y \rightarrow X$. Consider for cycle $i$, we have task specification $t$, original data $s_0$, current data $s_i$, current input $x_i = t + s_i$, and current output $y_i = f(x_{i-1})$. The discriminator compares $s_0$ and $s_i$ to produce a hint, i.e. $d(s_0, s_i, y_i) \rightarrow hint_i$. This hint is used for two updates: we update current data $s_{i+1} = s_i + hint_i$, and update current input $x_{i+1} = t + s_{i+1}$.

The iterative refinement process is performed $N$ times, or until $s_0$ and $s_i$ are consistent. This is formalized as:
\begin{align}
    x_0 &= t + s_0 \\
    y_i &= f(x_{i-1}) = f(t + s_{i-1}) \\
    s_i &= g(y_i)  \\
    x_i &= t + s_i \\
    hint_{i+1} &= d(s_0, s_i, y_i) \\
    s_{i+1} &= s_i + hint_{i+1} \\
    x_{i+1} &= t + s_{i+1}
\end{align}

Algorithm~\ref{alg:cycleprompt} demonstrates the approach in pseudocode.
   
\begin{algorithm}[tb]
\caption{CyclePrompt}
\label{alg:cycleprompt}
\begin{algorithmic}
\STATE {\bfseries Input:} original data $s_0$, task specification $t$, input $x_i=t + s_i$ number of cycles $N$, forward generator $f$, backward generator $g$, discriminator $d$\\
\STATE Apply initial cycle $g(f(x_0)) = g(y_1) = x_1$.
\FOR{$i=1$ {\bfseries to} $N-1$}
\STATE Apply $f$ to $x_{i-1}$ to get $y_i = f(x_{i-1})$
\STATE Apply $g$ to $y_i$ to get $g(f(x_i)) = s_i$
\STATE Apply $d$ to $(s_0,s_i,y_i)$ to get $hint_{i+1}$
\STATE Update $x_{i+1} = t + s_i + hint_{i+1}$
\IF{$\texttt{Consistent}(s_0, s_{i+1})$}
\STATE Break
\ENDIF
\ENDFOR
\end{algorithmic}
\end{algorithm}

\section{Datasets}  
   
The selection of datasets for our experiments was guided by the need to test our approach across different domains and tasks. We chose three datasets: HumanEval, VQAv2, and FigureQA. These datasets cover a range of tasks, including code generation and vision-language tasks, allowing us to evaluate the versatility and effectiveness of our method.  
   
\subsection{HumanEval}  
   
HumanEval \cite{chen2021evaluating} is a dataset created by OpenAI for the purpose of evaluating the performance of AI models in generating code. It consists of a series of tasks that require the model to generate Python code to solve a given problem. The tasks are designed to be solvable by a competent programmer in a few minutes and cover a wide range of topics, including string manipulation, data structures, algorithms, and more. The dataset was created as part of the HumanEval competition, where models are ranked based on their performance on these tasks.  
   
In our experiments, we use the HumanEval dataset to test our approach on code generation tasks. We use the tasks in the dataset as inputs to our model, and evaluate the quality of our generated code using the evaluation methods provided by the benchmark authors.
   
\subsection{VQAv2}  
   
The Visual Question Answering version 2 (VQAv2) dataset \cite{balanced_vqa_v2} is a large-scale dataset for vision-language tasks. It contains open-ended questions about images, which require an understanding of the content of the image to answer. The dataset was created to improve upon the limitations of the original VQA dataset, with a particular focus on reducing biases in the data.  
   
In our experiments, we use the VQAv2 dataset to test our approach on vision-language tasks. We use the images in the dataset as inputs to our model, and evaluate the quality of our generated captions, using the ground truth question-answer pairs provided by the authors.
   
\subsection{FigureQA}  
   
FigureQA \cite{kahou2017figureqa} is a dataset created by Microsoft Research for the task of visual reasoning on diagrammatic and statistical images. It contains questions and answers about synthetically generated figures, including bar graphs, line graphs, and pie charts. The dataset was created to push the boundaries of what models can achieve in the realm of visual reasoning, with a focus on understanding and interpreting visual data.  
   
In our experiments, we use the FigureQA dataset to further test our approach on vision-language tasks, knowing that models sometimes perform worse in subdomains that are different from natural images. Similar to with VQAv2, we evaluate the quality of our generated captions using the ground truth question-answer pairs provided by the authors.

\section{Experiment: Code Generation}

To perform code generation, we run CyclePrompt on the Text-Code-Text cycle. Here, we begin with a language description of our coding task, and translate it to Python code (this is the forward generator). Then, we describe that generated code, yielding a new description (this is the backward generator). Finally, we observe three things -- original description, new description, and generated code -- and by comparing the two descriptions, we determine a better task description \textit{that would produce code whose description matches the original} (this is the discriminator). Intuitively, if the description of our generated code matches our original task description, then we have accomplished the task faithfully. This cycle-consistency can be done entirely within a single model (GPT4, in our case), and yields precise code and descriptions after several cycles.

CyclePrompt performs well on code generation tasks by exploiting two important dynamics: (1) the model has good understanding of code (in this case Python), and (2) the code generation step is sensitive to precise changes described in language. Together, this means that refinements to task descriptions at each cycle do actually improve code generation in subsequent rounds. We will see that such sensitivity does not exist in the vision-language space, which hampers performance there.

Table~\ref{table:code-results} demonstrates the results on the HumanEval code generation benchmark. Our CyclePrompt method reaches third place overall on the leaderboard, with accuracy of 87.2\%, beaten only by methods that utilize outside tooling. Among methods that are purely prompt-based, without the use of external tools, our method is first.

We provide exact prompts used in Appendix~\ref{sec:appendix-code-generation}.

\begin{table}[t]  
\caption{Code Generation Results on HumanEval Benchmark}  
\label{table:code-results}  
\vskip 0.15in  
\begin{center}  
\begin{small}  
\begin{sc}  
\begin{tabular}{lcc}  
\toprule  
Method & Acc. & Support Infrastructure \\  
\midrule
GPT4 & 80.5 &  -- \\
Parsel & 85.1 & Codex, Constraint Solver  \\
MetaGPT & 85.9 & Interpreter, Web, PubSub  \\
ANPL & 86.6 & User Interaction \\
OctoPack & 86.6 &  Fine-Tune \\
\textbf{CP (ours)} & \textbf{87.2} & -- \\
Reflexion & 91.0 &  Compiler, Interpreter\\
LATS & 94.4 & MCTS  \\
\bottomrule  
\end{tabular}  
\end{sc}  
\end{small}  
\end{center}    
\vskip -0.1in  
\end{table}

\section{Experiment: Image Captioning}

To perform image captioning, we run CyclePrompt on the
Image-Text-Image cycle. Here, we begin with an image input, and generate a caption for it (this is the forward generator). Then, we use that caption to generate a new image (this is the backward generator). Finally, we observe three things -- original image, new image, and generated caption -- and by comparing the two images, we create an updated caption [and subsequently generated image], \textit{whose caption would generate an image that matches the original} (this is the discriminator). If these generated images match our original image, then we have accomplished the task faithfully.

We share examples of discriminator input and image generation outputs for VQAv2 and FigureQA in Fig.~\ref{fig:io-vqav2} and Fig.~\ref{fig:io-figureqa}, respectively, and provide corresponding caption generation outputs in Appendix~\ref{appendix-sample-cycle-descriptions}.

\subsection{Evaluation metrics}

To test whether our generated captions are useful, we measure accuracy on well-known question-answer (QA) datasets.
As an upper limit of performance, we compute QA accuracy of a VQA model which see the image itself, and then we measure for our approach, i.e. QA accuracy given only generated captions. We also employ the DA-Score method of \cite{balanced_vqa_v2} to measure vision-language alignment.

In summary, considering original image $I$, caption $C$, question $Q$, and answer $A$, we measure with the following tools, for various settings:
\begin{align}
    \text{Visual QA Model}(I, Q) &\rightarrow A \\ \nonumber
    \text{Text QA Model}(C, Q) &\rightarrow A \\ \nonumber
    \text{DA-Score}(I, Q) &\rightarrow A \nonumber
\end{align}

Our suite of metrics aims to measure whether our generated captions are (1) as useful for question-answering as the original image, and (2) correct in content and details.

\textbf{Visual QA} is used to test how well a model can answer question based on the original image itself. This is considered the  ``gold standard" performance, since the model has full information. We use GPT4V as a VQA model, with image and question as input, and generate an answer. The answer is then compared to the ground truth answer.

\textbf{Text QA} is our primary metric, focusing on the question-answering ability of a piece of text.
We want to know: ``Can the CyclePrompt-generated caption provide as much information for question-answering as the original image?" For this case, we use GPT4 as a text QA model, \textit{with caption and question as input}, and generate an answer. The answer is then compared to the ground truth answer. Here, we seek a comparable baseline, and use captions provided in a zero-shot manner from GPT4V, when asked to produce a detailed caption. 

\textbf{DA-Score:} The metrics discussed so far are unidirectional, i.e. they check whether information from the image exists in the caption. But what if our candidate caption introduces spurious details? To account for this, we use the  Decompositional-Alignment-Score \cite{balanced_vqa_v2}. This score is given a complex text, and decomposes it into a set of disjoint assertions. The alignment of each assertion with images is then measured using a BLIP VQA model. Finally, alignment scores for different assertions are combined a posteriori to give the final text-to-image alignment score. 

\subsection{Experimental setup}

We use GPT4V as our forward generator to map images to text, and DALL$\cdot$E 3 as our backward generator to map text to images. To compare images in our discriminator function, we use an independent instance of GPT4V. For each dataset, we fix 200 random samples for all experimental settings. We find this subsample size large enough to contain diverse examples, while small enough to accommodate compute time and budget. In Table~\ref{table:caption-results}, the methods compared are: (1) \textit{GPT4V(image)} where GPT4V answers benchmark questions directly from the image, (2) \textit{GPT4V(our caption)} where accuracy and DA-Score are computed for questions generated from CyclePrompt captions, (3) \textit{GPT4V(0-shot caption)} where accuracy and DA-Score are computed for questions generated by zero-shot GPT4V captions, and (4) \textit{DA-baseline} where DA-Score is evaluated based on image and benchmark questions.

We provide exact prompts used in Appendix~\ref{sec:appendix-vision-language}.


\subsection{Captioning Performance}

Table~\ref{table:caption-results} includes performance metrics for CyclePrompt, compared to various baselines. $ACC_{bench}$ is accuracy on benchmark questions, and DA-score measures image-caption alignment.

In VQAv2 results, GPT4V(image) performs best, with accuracy value of 0.820 for benchmark questions. This means that 82\% of questions were answered correctly, when given the image and benchmark question directly.
As expected, accuracy is lower when only captions are available, but CyclePrompt captions perform on par with or better than zero-shot GPT4V captions, when comparing accuracy (0.652 versus 0.632) and DA-Scores (0.682 versus 0.699 [positive assertions only] and 0.479 versus 0.509 [negative assertions included]).

For the diagrammatic images in FigureQA, GPT4V also gives highest accuracy (0.584), but again CyclePrompt achieves higher accuracy compared to zero-shot GPT4V (0.512 versus 0.477). 
We believe that CyclePrompt allows the model to collect details that would not have been described in a single pass. DA-Scores for CyclePrompt (0.557, 0.237) are also higher than those of zero-shot GPT4V (0.553, 0.231) indicating that the assertions generated from CyclePrompt captions were more aligned to the original image than assertions generated from zero-shot GPT4V captions. The DA-Scores for DA-baseline (0.637, 0.391) are the highest, possibly due to the relative simplicity of benchmark questions, compared to questions from highly-detailed captions, which often contained technical or aesthetic details.

\begin{figure}[ht]  
\centering  
\begin{subfigure} 
  \centering  
  \includegraphics[width=.8\linewidth]{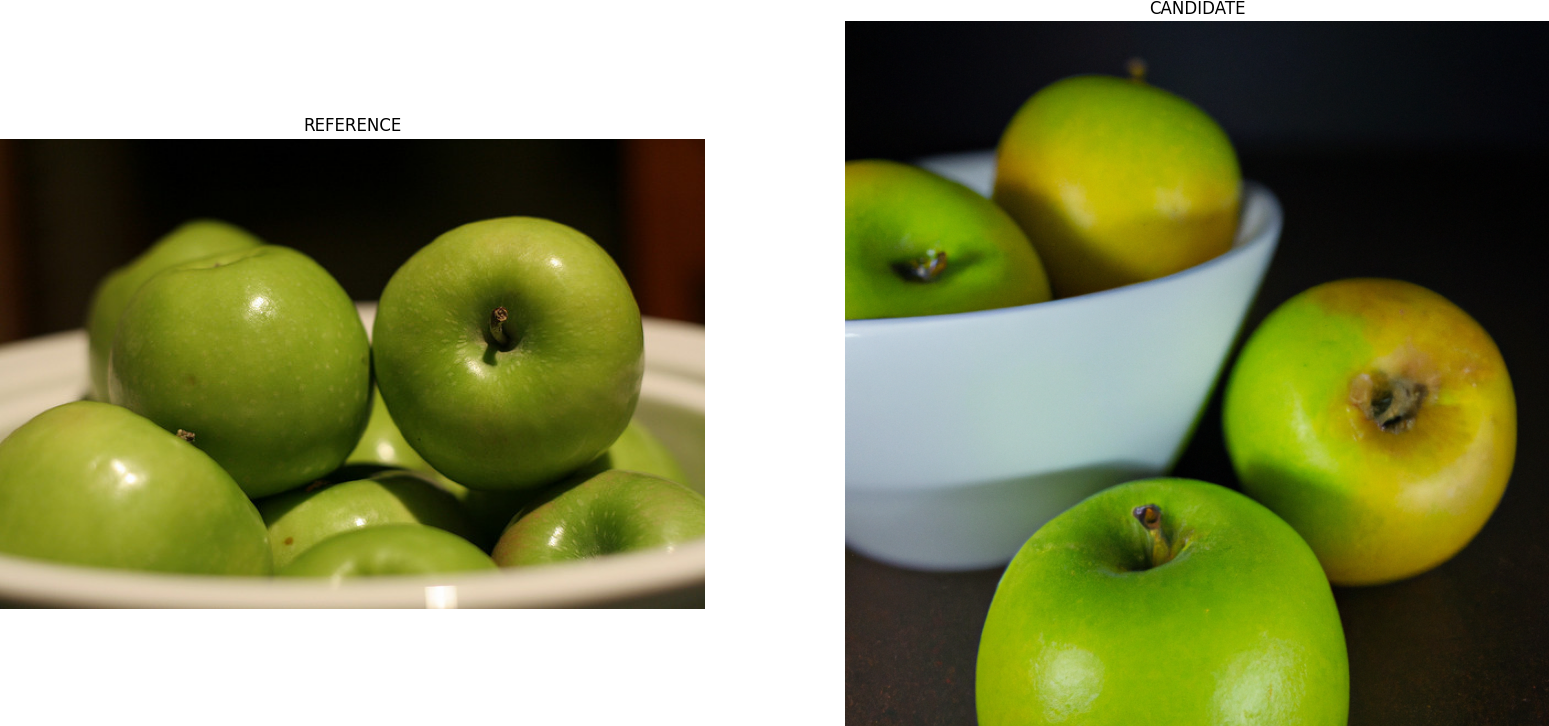}  
  \caption{Composite input to discriminator}  
  \label{fig:composite-vqav2}  
\end{subfigure}  
\begin{subfigure}  
  \centering  
  \includegraphics[width=.8\linewidth]{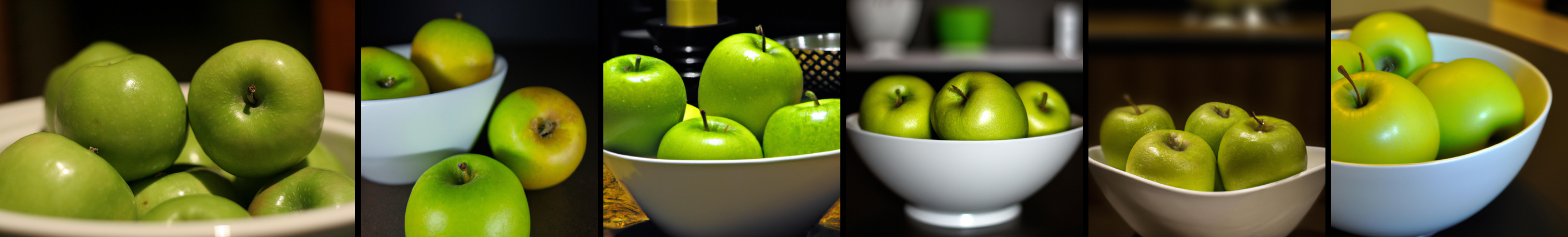}  
  \caption{Original image (left) and subsequent cycle generations (right).}  
  \label{fig:generations-vqav2}  
\end{subfigure}  
\caption{CyclePrompt inputs and outputs. \underline{Final caption:} \textit{``Several bright green apples with a smooth, shiny texture are placed in a white bowl with a wide rim. The apples are unblemished, except for one in the foreground that has a small, dark indentation near the stem. The bowl sits on a dark surface, and the background is a blurred, dark brown, providing a stark contrast to the vibrant green of the apples. The apples are closely packed together, with one apple prominently in the foreground, slightly obscuring the apples behind it. The lighting is soft and diffused, highlighting the apples' texture and color. The apples have visible white speckles, and the bowl has a subtle shadow cast on the right side. The apples appear more matte than glossy, and the bowl's rim is thick and slightly curved outward."}}  
\label{fig:io-vqav2}  
\end{figure}  

\begin{figure}[ht]  
\centering  
\begin{subfigure}
  \centering  
  \includegraphics[width=.8\linewidth]{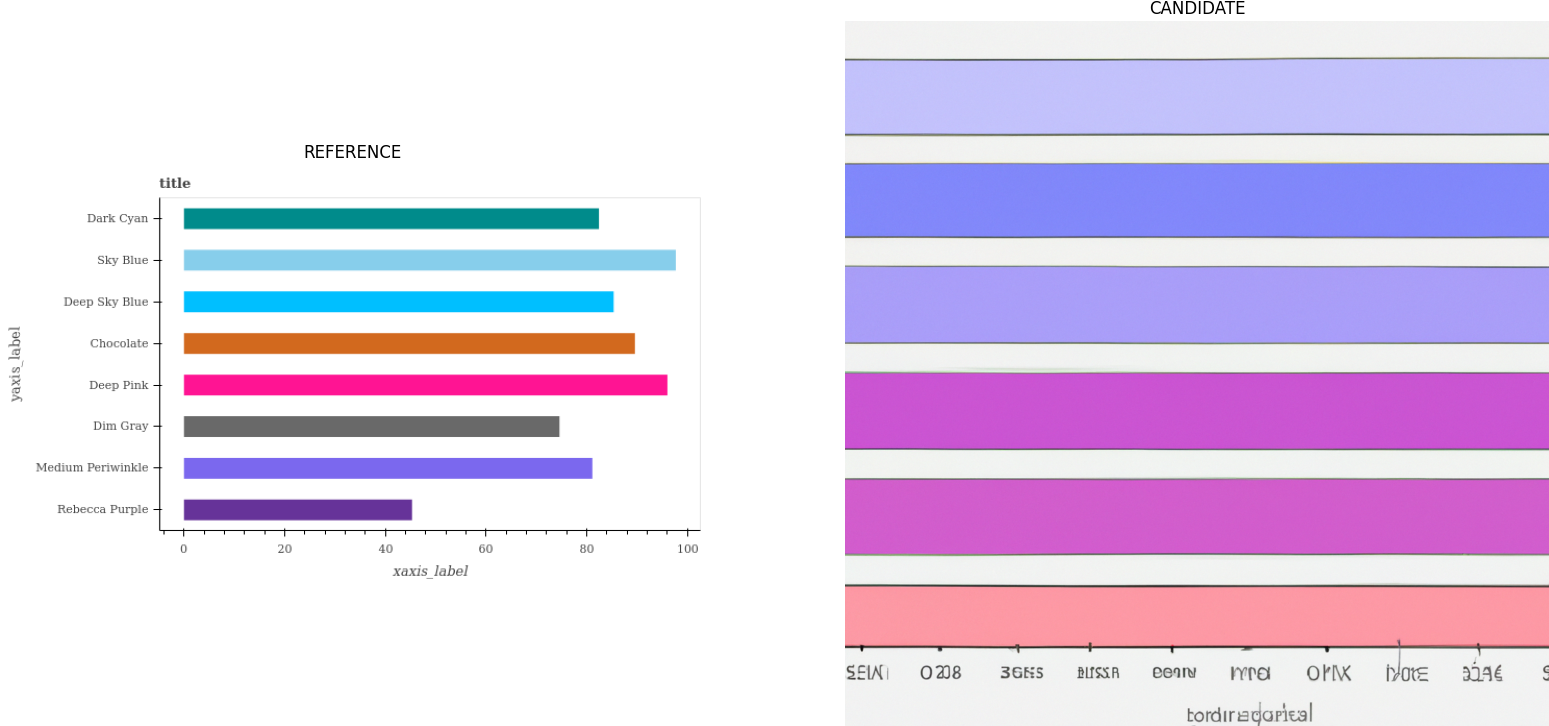}  
  \caption{Composite input to discriminator}  
  \label{fig:composite-figureqa}  
\end{subfigure}  
\begin{subfigure} 
  \centering  
  \includegraphics[width=.8\linewidth]{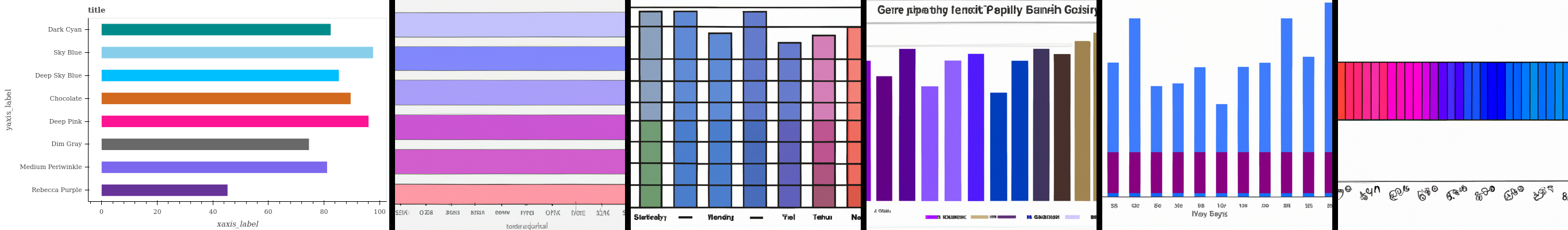}  
  \caption{Original image (left) and subsequent cycle generations (right).}  
  \label{fig:generations-figureqa}  
\end{subfigure}  
\caption{CyclePrompt inputs and outputs. \underline{Final caption:} \textit{``A horizontal bar graph with eight bars in distinct colors, each labeled with a color name on the left side. The bars are arranged from top to bottom in the order of Dark Cyan, Sky Blue, Deep Sky Blue, Chocolate, Deep Pink, Dim Gray, Medium Periwinkle, and Rebecca Purple. The x-axis is labeled `xaxis label' with a scale from 0 to 100, and the y-axis is labeled `yaxis label.' The graph has a title at the top that reads `title.' The bars have varying lengths representing different values on the x-axis, with Dark Cyan being the longest and Rebecca Purple being the shortest. The graph is a clear, 2D representation with no grid lines, and the bars are solid with no patterns or textures. The title, axis labels, and color labels are all clearly legible. The graph background is white, and the bars are not stacked."}}  
\label{fig:io-figureqa}  
\end{figure}

\begin{table}[t]  
\caption{Comparison of VQA Accuracies}  
\label{table:caption-results}  
\vskip 0.15in  
\begin{center}  
\begin{small}  
\begin{sc}  
\begin{tabular}{lcc}  
\toprule
Method & $Acc_{bench}$ & DA-Score (p, n) \\  
\midrule  
GPT4V(image) & 0.820 & -- \\  
GPT4(our caption) & 0.652 & 0.682, 0.479 \\  
GPT4(0-shot caption) & 0.632 & 0.699, 0.509 \\
da-baseline & -- & 0.467, 0.102 \\
\bottomrule  
\end{tabular}  
\end{sc}  
\end{small}  
\end{center}  
\centering{(a) VQAv2}  
\vskip 0.15in  
\begin{center}  
\begin{small}  
\begin{sc}  
\begin{tabular}{lcc}  
\toprule
Method & $Acc_{bench}$ & DA-Score (p, n) \\  
\midrule  
GPT4V(image) & 0.584* & -- \\  
GPT4(our caption) & 0.512* & 0.557, 0.237 \\  
GPT4(0-shot caption) & 0.477* & 0.553, 0.231 \\
da-baseline & -- & 0.637, 0.391 \\
\bottomrule  
\end{tabular}  
\end{sc}  
\end{small}  
\end{center}  
\centering{(b) FigureQA. *Reported on subset, due to compute time constraints.}  
\vskip -0.1in  
\end{table}  

\section{Case Study: Image Generation}

We observe different dynamics in the image generation case, i.e. the Text-Image-Text cycle. This is due to the relatively higher complexity of the target image space, making the cycle unbounded. As a reminder of the opposite, recall the image captioning case. There, our input is an image, which is higher-complexity and detail-rich; and our output is a caption, which is lower-complexity and naturally summarizing. This dynamic is stable, because generated captions are always less detailed than the image. In contrast, when generating images from text, generated images continuously introduce new details, all of which can be consistent with the original text.

\begin{figure*}[ht]  
\centering  
\begin{subfigure} 
  \centering  
  \includegraphics[width=\linewidth]{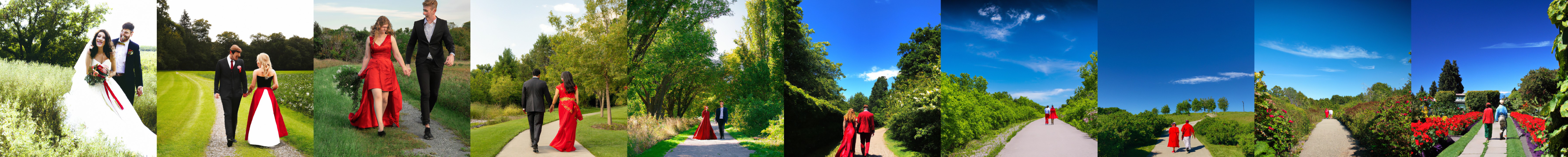}  
  \caption{Text-Image-Text cycle for ``A happy day"}  
  \label{fig:text-image-text-a-happy-day}  
\end{subfigure}  
\begin{subfigure} 
  \centering  
  \includegraphics[width=\linewidth]{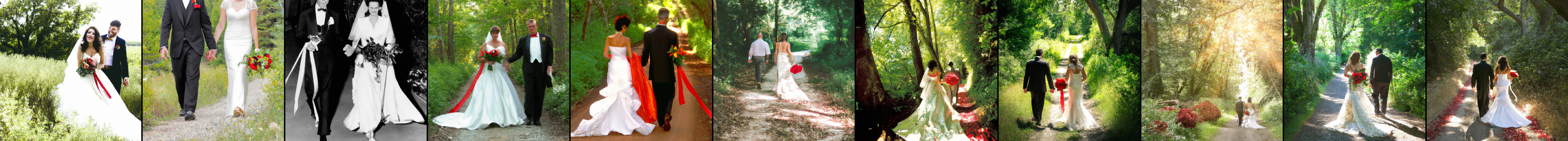}  
  \caption{Image-Text-Image cycle on first generation of \ref{fig:text-image-text-a-happy-day}.}  
  \label{fig:image-text-image-a-happy-day}  
\end{subfigure}
\begin{subfigure} 
  \centering  
  \includegraphics[width=\linewidth]{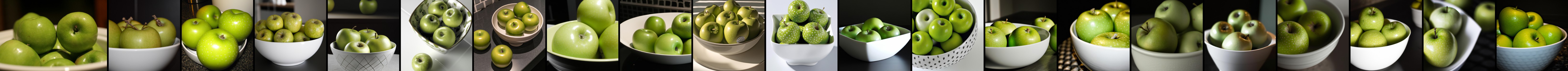}  
  \caption{Twenty cycles in Image-Text-Image.}  
  \label{fig:image-text-image-apples-20-cycles}  
\end{subfigure}
\caption{Comparison of Text-Image-Text (image generation) to Image-Text-Image (image captioning). When the input space is lower-complexity (e.g. text to image, in \ref{fig:text-image-text-a-happy-day}), the output space can comply with the input while continuously changing. When the input space is higher-complexity (e.g. image to text, in \ref{fig:image-text-image-a-happy-day},\ref{fig:image-text-image-apples-20-cycles}), both spaces are constrained and converge.}  
\label{fig:cycle-direction-comparison}  
\end{figure*}

Figure~\ref{fig:cycle-direction-comparison} provides an extreme example to illustrate. If we begin from text prompt ``a happy day", and perform a cycle by generating an image and then captioning it; then our updated caption will express details introduced by our image generator. Normally, this is acceptable, but because text is naturally lower-complexity than images, the image generator continues to introduce new details, while still complying with the original input text. This is shown in Fig.~\ref{fig:text-image-text-a-happy-day}, where generated images change gradually as the cycles progress (from wedding photoshoot to generic garden walk), all while satisfying the original input ``a happy day".

To demonstrate the inverse relationship of complexity (where we would seek to use CyclePrompt), we show in Fig.~\ref{fig:image-text-image-a-happy-day} what happens if we start the cycle with an image. Here, we think of the image as \textit{more prescriptive} than text, thereby keeping even long sequences of cycles constrained and converged. Figure~\ref{fig:image-text-image-apples-20-cycles} provides another example of this convergence, for 20 cycles.

\section{Discussion}

CyclePrompt can give state-of-the-art performance without external tooling, but its utility obviously depends on the strength of the forward, backward, and discriminator functions.

We find that the forward generator is most important, followed by the discriminator, and finally the backward generator.

The forward direction is most critical because without a strong forward generator, the system does not have the ability to execute on feedback. In the Text-Code-Text cycle, we see that our model (GPT4) is extremely responsive to changes in our input text, and it faithfully translates text instructions into code. In the Image-Text-Image cycle, we similarly see that our model (GPT4V) reliably provides accurate descriptions for images. In short, without a good forward generator, the system could not improve, even with a good signal.

The discriminator is of secondary importance, because it defines the upper limit of how much refinement CyclePrompt can deliver. Because instructions are pass via text, we benefit from the excellent language understanding of our model.

Finally, a strong backward generator is indeed helpful, but CyclePrompt can still produce useful results when it is weak. This is most evident in the Image-Text-Image cycle, where image generation is weaker, i.e. generated images frequently do \textit{not} comply with highly detailed captions. Even still, the fact that it samples in a neighboring semantic space, paired with a good discriminator, can inform how to more precisely caption the original image.

\section{Related Work}

The field of LLMs has made significant strides in recent years \cite{team2023gemini, brown2020language, openai2023gpt4, yang2023harnessing, touvron2023llama, gunasekar2023textbooks, anil2023palm, wu2023autogen}. 
These advancements not only expanded the application of LLMs, but also highlighted the importance of effective interaction with these models. Substantial research has focused on prompt engineering
to elicit specific responses from Large Language Models \cite{wei2022chain, kaddour2023challenges}, and numerous approaches to prompt engineering have been explored, including context learning, chain-of-thought, and multi-turn prompting techniques \cite{wei2022chain, schick2020s, yao2023tree, yao2022react, madaan2023self}. These strategies enhance our interaction with LLMs, enabling users to obtain more accurate and relevant responses from models.

In the subfield of context learning, few-shot examples in the prompt were shown to enhance model performance \cite{brown2020language}. But the technique is not flawless, particularly when addressing more intricate reasoning tasks. To address this, \cite{wei2022chain} proposed Chain-of-Thought Prompting (CoT), which provides LLMs with step-by-step reasoning examples instead of standard question-answer examples. This helps models to generate a reasoning path that breaks down complex reasoning into multiple, more manageable steps \cite{kojima2022large, wei2022chain}.

To further enhance CoT Prompting, \cite{wang2022self} introduced a decoding strategy called self-consistency, which replaces the naive greedy decoding used in CoT Prompting. Self-consistency employs a set of manually written chain-of-thought examples, and samples a collection of candidate outputs from the language model's decoder. Finally, answers are combined by considering the sampled reasoning paths and choosing the most consistent answer among the generated options \cite{wang2022self}.
Another related work that expands CoT prompting technique is the Tree of Thoughts (ToT) framework \cite{yao2023tree}, which facilitates the investigation of cohesive text segments, known as `thoughts'. These thoughts act as intermediary stages in the process of solving problems.

Another work that enhances CoT reasoning is ReAct\cite{yao2022react}. ReAct integrates reasoning and acting with language models to address language reasoning and decision-making tasks. This method enables the model to carry out dynamic reasoning for creating, maintaining, and adjusting high-level action plans, while interacting with external environments such as Wikipedia to incorporate supplementary information. This is an improvement over the CoT reasoning, which is a static black box and relies on its internal representations to generate thoughts and is not grounded in the external world.


In the space of translation, Iterative Domain-Repaired Back-Translation \cite{wei2020iterative} focuses on addressing the challenges of domain-specific translation in low-resource settings by utilizing in-domain monolingual data through the back-translation method. The authors propose a framework that incorporates a Domain-Repair (DR) model to enhance the quality of translations in synthetic bilingual data. The DR model is trained on round-trip translated monolingual sentences and aims to jointly optimize both the DR and Neural Machine Translation (NMT) models. This innovative approach demonstrates the potential of the DR model in refining translations and improving domain adaptation in scenarios with limited resources.

There has also been significant work in ``prompt learning" (also known as ``prompt tuning", ``context optimization", and ``soft prompting") where optimization is performed in the embedding space of the prompt, rather than in the semantic space \cite{wei2023improving,zhu2023prompt,zhou2022conditional,gu2023systematic}. These methods require additional training, as well as access to internals of vision-language models, to pass in custom prompt embeddings. Our method, by contrast, is an exclusive black-box model user.

Most relevant to our work is SELF-REFINE \cite{madaan2023self}, which also employs iterative prompting. This technique aims to improve the initial outputs of LLMs by employing ongoing feedback and refinement processes. The fundamental idea consists of generating an initial output with an LLM, then having the same LLM provide feedback on its output and iteratively use this feedback for self-enhancement. SELF-REFINE performs in-context learning with few-shot examples of (input, output, feedback) triples, in order to predict feedback for refinement. Our work differs in two important ways: (1) Our approach requires no training data or expertise to create in-context examples, and (2) our approach is well-suited for multimodal applications, where a map [and its inverse] are commonly developed between a new modality and language, thus enabling the desired cycle-supervision.

Finally, the notion of feedback as a ``semantic" or textual gradient is also explored in \cite{pryzant2023automatic} and \cite{shinn2023reflexion}, but both require ground truth data or an external environment to score the value of potential changes.





\section{Future Work}
Our work with CyclePrompt has opened several paths for future research.

First, the method can be compared to and combined with multi-agent systems with access to external environments. A hybrid of these approaches might combine their strengths to produce a problem-solving system that is both efficient and high-performing.

Second, we note that the success of CyclePrompt is tied to specific dynamics, including the quality of the forward function, the precision of the discriminator, and to a lesser extent, the quality of the backward function. Future research could further investigate these dynamics to characterize modality \textit{misalignment}; and recognition of such asymmetries in sensitivity, precision, and complexity could lead to improved understanding of model knowledge.

Finally, an intriguing question is how to formalize the notion of ``semantic gradient descent". What are the necessary conditions for the discriminator to enable long-range optimization in the semantic space? What stopping criteria, limits, or convergence properties can be derived? Machine learning models perform millions of updates to numerical parameters. 
Can thousands or millions of semantic updates be run, instead of less than ten, as is typical in the existing research? Can monotonic improvement be guaranteed?

We believe this work provides a starting point for exploring these topics, demonstrating both interesting open questions about foundation model knowledge, as well as practical strategies for extracting state-of-the-art performance from existing systems.

\bibliography{references}

\begin{thebibliography}{10}

\bibitem{shinn2023reflexion}
Noah Shinn, Federico Cassano, Ashwin Gopinath, Karthik~R Narasimhan, and Shunyu Yao.
\newblock Reflexion: Language agents with verbal reinforcement learning.
\newblock In {\em Thirty-seventh Conference on Neural Information Processing Systems}, 2023.

\bibitem{zhu2017unpaired}
Jun-Yan Zhu, Taesung Park, Phillip Isola, and Alexei~A Efros.
\newblock Unpaired image-to-image translation using cycle-consistent adversarial networks.
\newblock In {\em Proceedings of the IEEE international conference on computer vision}, pages 2223--2232, 2017.

\bibitem{wang2019learning}
Xiaolong Wang, Allan Jabri, and Alexei~A Efros.
\newblock Learning correspondence from the cycle-consistency of time.
\newblock In {\em Proceedings of the IEEE/CVF Conference on Computer Vision and Pattern Recognition}, pages 2566--2576, 2019.

\bibitem{wang2022self}
Xuezhi Wang, Jason Wei, Dale Schuurmans, Quoc Le, Ed~Chi, Sharan Narang, Aakanksha Chowdhery, and Denny Zhou.
\newblock Self-consistency improves chain of thought reasoning in language models.
\newblock {\em arXiv preprint arXiv:2203.11171}, 2022.

\bibitem{yao2022react}
Shunyu Yao, Jeffrey Zhao, Dian Yu, Nan Du, Izhak Shafran, Karthik Narasimhan, and Yuan Cao.
\newblock React: Synergizing reasoning and acting in language models.
\newblock {\em arXiv preprint arXiv:2210.03629}, 2022.

\bibitem{yao2023tree}
Shunyu Yao, Dian Yu, Jeffrey Zhao, Izhak Shafran, Thomas~L Griffiths, Yuan Cao, and Karthik Narasimhan.
\newblock Tree of thoughts: Deliberate problem solving with large language models.
\newblock {\em arXiv preprint arXiv:2305.10601}, 2023.

\bibitem{hao2023reasoning}
Shibo Hao, Yi~Gu, Haodi Ma, Joshua~Jiahua Hong, Zhen Wang, Daisy~Zhe Wang, and Zhiting Hu.
\newblock Reasoning with language model is planning with world model.
\newblock {\em arXiv preprint arXiv:2305.14992}, 2023.

\bibitem{madaan2023self}
Aman Madaan, Niket Tandon, Prakhar Gupta, Skyler Hallinan, Luyu Gao, Sarah Wiegreffe, Uri Alon, Nouha Dziri, Shrimai Prabhumoye, Yiming Yang, et~al.
\newblock Self-refine: Iterative refinement with self-feedback.
\newblock {\em arXiv preprint arXiv:2303.17651}, 2023.

\bibitem{pryzant2023automatic}
Reid Pryzant, Dan Iter, Jerry Li, Yin~Tat Lee, Chenguang Zhu, and Michael Zeng.
\newblock Automatic prompt optimization with" gradient descent" and beam search.
\newblock {\em arXiv preprint arXiv:2305.03495}, 2023.

\bibitem{zhou2023language}
Andy Zhou, Kai Yan, Michal Shlapentokh-Rothman, Haohan Wang, and Yu-Xiong Wang.
\newblock Language agent tree search unifies reasoning acting and planning in language models.
\newblock {\em arXiv preprint arXiv:2310.04406}, 2023.

\bibitem{chen2021evaluating}
Mark Chen, Jerry Tworek, Heewoo Jun, Qiming Yuan, Henrique Ponde de~Oliveira Pinto, Jared Kaplan, Harri Edwards, Yuri Burda, Nicholas Joseph, Greg Brockman, et~al.
\newblock Evaluating large language models trained on code.
\newblock {\em arXiv preprint arXiv:2107.03374}, 2021.

\bibitem{balanced_vqa_v2}
Yash Goyal, Tejas Khot, Douglas Summers{-}Stay, Dhruv Batra, and Devi Parikh.
\newblock Making the {V} in {VQA} matter: Elevating the role of image understanding in {V}isual {Q}uestion {A}nswering.
\newblock In {\em Conference on Computer Vision and Pattern Recognition (CVPR)}, 2017.

\bibitem{kahou2017figureqa}
Samira~Ebrahimi Kahou, Vincent Michalski, Adam Atkinson, {\'A}kos K{\'a}d{\'a}r, Adam Trischler, and Yoshua Bengio.
\newblock Figureqa: An annotated figure dataset for visual reasoning.
\newblock {\em arXiv preprint arXiv:1710.07300}, 2017.

\bibitem{team2023gemini}
Gemini Team, Rohan Anil, Sebastian Borgeaud, Yonghui Wu, Jean-Baptiste Alayrac, Jiahui Yu, Radu Soricut, Johan Schalkwyk, Andrew~M Dai, Anja Hauth, et~al.
\newblock Gemini: a family of highly capable multimodal models.
\newblock {\em arXiv preprint arXiv:2312.11805}, 2023.

\bibitem{brown2020language}
Tom Brown, Benjamin Mann, Nick Ryder, Melanie Subbiah, Jared~D Kaplan, Prafulla Dhariwal, Arvind Neelakantan, Pranav Shyam, Girish Sastry, Amanda Askell, et~al.
\newblock Language models are few-shot learners.
\newblock {\em Advances in neural information processing systems}, 33:1877--1901, 2020.

\bibitem{openai2023gpt4}
OpenAI.
\newblock Gpt-4 technical report, 2023.

\bibitem{yang2023harnessing}
Jingfeng Yang, Hongye Jin, Ruixiang Tang, Xiaotian Han, Qizhang Feng, Haoming Jiang, Bing Yin, and Xia Hu.
\newblock Harnessing the power of llms in practice: A survey on chatgpt and beyond.
\newblock {\em arXiv preprint arXiv:2304.13712}, 2023.

\bibitem{touvron2023llama}
Hugo Touvron, Louis Martin, Kevin Stone, Peter Albert, Amjad Almahairi, Yasmine Babaei, Nikolay Bashlykov, Soumya Batra, Prajjwal Bhargava, Shruti Bhosale, et~al.
\newblock Llama 2: Open foundation and fine-tuned chat models.
\newblock {\em arXiv preprint arXiv:2307.09288}, 2023.

\bibitem{gunasekar2023textbooks}
Suriya Gunasekar, Yi~Zhang, Jyoti Aneja, Caio C{\'e}sar~Teodoro Mendes, Allie Del~Giorno, Sivakanth Gopi, Mojan Javaheripi, Piero Kauffmann, Gustavo de~Rosa, Olli Saarikivi, et~al.
\newblock Textbooks are all you need.
\newblock {\em arXiv preprint arXiv:2306.11644}, 2023.

\bibitem{anil2023palm}
Rohan Anil, Andrew~M Dai, Orhan Firat, Melvin Johnson, Dmitry Lepikhin, Alexandre Passos, Siamak Shakeri, Emanuel Taropa, Paige Bailey, Zhifeng Chen, et~al.
\newblock Palm 2 technical report.
\newblock {\em arXiv preprint arXiv:2305.10403}, 2023.

\bibitem{wu2023autogen}
Qingyun Wu, Gagan Bansal, Jieyu Zhang, Yiran Wu, Shaokun Zhang, Erkang Zhu, Beibin Li, Li~Jiang, Xiaoyun Zhang, and Chi Wang.
\newblock Autogen: Enabling next-gen llm applications via multi-agent conversation framework.
\newblock {\em arXiv preprint arXiv:2308.08155}, 2023.

\bibitem{wei2022chain}
Jason Wei, Xuezhi Wang, Dale Schuurmans, Maarten Bosma, Fei Xia, Ed~Chi, Quoc~V Le, Denny Zhou, et~al.
\newblock Chain-of-thought prompting elicits reasoning in large language models.
\newblock {\em Advances in Neural Information Processing Systems}, 35:24824--24837, 2022.

\bibitem{kaddour2023challenges}
Jean Kaddour, Joshua Harris, Maximilian Mozes, Herbie Bradley, Roberta Raileanu, and Robert McHardy.
\newblock Challenges and applications of large language models.
\newblock {\em arXiv preprint arXiv:2307.10169}, 2023.

\bibitem{schick2020s}
Timo Schick and Hinrich Sch{\"u}tze.
\newblock It's not just size that matters: Small language models are also few-shot learners.
\newblock {\em arXiv preprint arXiv:2009.07118}, 2020.

\bibitem{kojima2022large}
Takeshi Kojima, Shixiang~Shane Gu, Machel Reid, Yutaka Matsuo, and Yusuke Iwasawa.
\newblock Large language models are zero-shot reasoners.
\newblock {\em Advances in neural information processing systems}, 35:22199--22213, 2022.

\bibitem{wei2020iterative}
Hao-Ran Wei, Zhirui Zhang, Boxing Chen, and Weihua Luo.
\newblock Iterative domain-repaired back-translation.
\newblock {\em arXiv preprint arXiv:2010.02473}, 2020.

\bibitem{wei2023improving}
Hongchen Wei and Zhenzhong Chen.
\newblock Improving generalization of image captioning with unsupervised prompt learning.
\newblock {\em arXiv preprint arXiv:2308.02862}, 2023.

\bibitem{zhu2023prompt}
Beier Zhu, Yulei Niu, Yucheng Han, Yue Wu, and Hanwang Zhang.
\newblock Prompt-aligned gradient for prompt tuning.
\newblock In {\em Proceedings of the IEEE/CVF International Conference on Computer Vision}, pages 15659--15669, 2023.

\bibitem{zhou2022conditional}
Kaiyang Zhou, Jingkang Yang, Chen~Change Loy, and Ziwei Liu.
\newblock Conditional prompt learning for vision-language models.
\newblock In {\em Proceedings of the IEEE/CVF Conference on Computer Vision and Pattern Recognition}, pages 16816--16825, 2022.

\bibitem{gu2023systematic}
Jindong Gu, Zhen Han, Shuo Chen, Ahmad Beirami, Bailan He, Gengyuan Zhang, Ruotong Liao, Yao Qin, Volker Tresp, and Philip Torr.
\newblock A systematic survey of prompt engineering on vision-language foundation models.
\newblock {\em arXiv preprint arXiv:2307.12980}, 2023.

\end{thebibliography}
\bibliographystyle{unsrt}

\newpage
\appendix
\onecolumn
\section{Code Generation}
\label{sec:appendix-code-generation}
Below are the prompts used for code generation.

\subsection{Forward Generator}
\begin{lstlisting}[language=]
Given the code below:

[code]

 please conclude and describe the task of the code.
\end{lstlisting}

\subsection{Backward Generator}
\begin{lstlisting}[language=]
You are a professional programmer. You will be given a coding task.

Please use python to write the code.

Your response should include only python code. no code comment, no description, no commentary, no docstring, just the python code.

Attention: NO code comment.
\end{lstlisting}

\subsection{Discriminator}
\begin{lstlisting}[language=]
We have two procedures, roughly corresponding to:
1. "go from task description to code", and
2. "go from code to task description".

We can achieve a cycle consistency, e.g. description -> generated code -> description, if the original task description and concluded task descriptions are equivalent. This cycle consistency can be achieved only if the generated code is correct. If the code is wrong, the concluded task description from the code will be different from original task description.
The original task description is: [Task description]
 
The generated code is: [code]
 
The concluded task description is: [Conclusion].
Our ultimate goal is to generate the correct code. Please try to find the potential errors/mistakes in the generated code, by observing and reflecting on the differences between the original task description and the concluded task description. Then advise on:
1. how to avoid potential mistakes/errors in the code;
2. how to simplify the code.
The entire response should focus on the specific advice to improve the code quality.
 
A few additional points:
 
(1) If you find inconsistency in the cycle, respond using the template below, where some example notes are provided in parentheses:

-----------------------

The original task is xxx (e.g., write a function counting from 0 to 10),
 
while the concluded task is xxx (e.g., write a function counting from 0 to infinity).
 
The difference is xxx (e.g., the range was changed).
 
The cause of the inconsistency is the generated code xxx (e.g., fail to set max value of range in for loop).
 
Therefore, my advice is:
 
xxx (e.g., ensure that the endpoints of the range are set correctly).

------------------------
 
 
(2) If you find that the cycle consistency has been achieved, respond using the template below:

-----------------------

The cycle is consistent, and I have no more advice.
\end{lstlisting}

\section{Vision-Language}
\label{sec:appendix-vision-language}

\subsection{Discriminator Function}

The following Python code snippet is the implementation of the discriminator or update function used in our CyclePrompt approach. This function, named update\_description, compares the original image and the generated image, and updates the description of the original image based on the differences observed.

This function is called during each cycle of the CyclePrompt process, specifically in the reflection and refinement step. It plays a crucial role in achieving cycle-consistency and improving the performance of the model.
\vspace{0.25cm}
\begin{lstlisting}[language=Python]
def update_description(  
    self, description, original_image_path, generated_image_path, save_dir, cycle_index  
):  
    """Compare the original image and the generated image."""  
    # Create composite image with original image on left and generated image on right  
    composite_image_path = os.path.join(save_dir, f"composite_{cycle_index}.png")  
    create_composite_image_plt(original_image_path, generated_image_path, composite_image_path)  

    # Define prompt to compare images and create new description  
    prompt = f"""  
        We are machine learning scientists, who are experimenting with cycle consistency in image generation.  

        The cycle we are testing is as follows. Given a reference image:  
        1. Generate a description of the reference image.  
        2. Use the description to generate a candidate image.  
        3. Compare the candidate image to the reference image.  
        4. Write an updated description of the reference image. The key to this step is that whatever differences are detected, they represent things that should be in the reference description, so that the new image is generated correctly (i.e. as close as possible to the reference).  
        5. Go back to step 2, and repeat the cycle.  

        We are currently doing steps 3 and 4, and we need your help.  

        The REFERENCE image is on the LEFT SIDE, and the CANDIDATE image is on the RIGHT.  
        The current description of the reference image is: {description}  

        Think about how the reference image is different from the candidate image, and write a new description of the reference image that takes into account those differences.  
          
        For example, suppose you have:  
          reference image: photo of black cat on brown leather sofa  
          candidate image: illustration of black cat on cloth sofa  
          description: "black cat on sofa"  
        Then the updated description would be something like:  
          "photograph of black cat on brown leather sofa"  
        because the reference image is a photograph, not an illustration, and the sofa is brown leather, not cloth.  
          
        Here are some tips on how to compare two images:  
            - Feature Correspondence: Do distinct features (edges, corners, textures, etc.) in one image correspond to the same features in the other image? If differences exist, describe the REFERENCE in terms of those features.  
            - Geometric Consistency: Do the spatial relationships between features within the images remain consistent. For example, if one image has large trees to the right of a tent, does the other image also have large trees to the right of the tent, or are the spatial relationships swapped or different? Or if the subject is facing one direction in one image, is it facing the same direction in the other image? If differences exist, describe the REFERENCE in terms of those relationships.  
            - Photometric Consistency: Do the images have consistent appearance in terms of lighting, color, and intensity. If differences exist, describe the REFERENCE in terms of those differences in appearance.  
            - Style Consistency: Are the image styles (photograph, painting, drawing, diagram, medical image) the same? If differences exist, describe the REFERENCE in terms of those differences in style.  
            - Semantic Consistency: Do objects and their parts maintain their identity and meaning across the images. For instance, a wheel of a bicycle should still be identifiable as a wheel of a bicyle in the other image. If differences exist, describe the REFERENCE in terms of those differences in semantics.  
            - Structural Integrity: Is the overall structure of the objects in the images preserved across the images? There should be no unnatural distortions or warping that compromise the object's recognizability. If differences exist, describe the REFERENCE in terms of those differences in structure.  
          
        NOTES:  
            - IMPORTANT: The above tips are useful for natural images. For graphical, statistical, or diagrammatic images, focus on the data itself and what kind of reasoning is being conveyed.  
            - Make sure to retain the major components or reasoning of the REFERENCE image.  
            - In the new description NEVER mention the reference or candidate images, i.e. DO NOT include a header or preamble like 'The reference image...' or 'The image on the left...'.  
            - ONLY output the new description. No other text, other than the new description.  
            - If the candidate image misses something, or contradicts the reference image in any of the ways described in the tips above (or otherwise), then EMPHASIZE this thing in the new reference description.  
            - Keep overall response to about 130 words or less. Feel free to shorter phrases or incomplete sentences, if it helps to include important details.  

        The new description of the reference image is:  
    """  

    # Run GPT4V on composite image, to get an updated description  
    updated_description = self.gpt4v_service.call(  
        composite_image_path, prompt, self.config["gpt4v_retries"], verbose=True  
    )  

    logging.info(f"\n\nCYCLE {cycle_index} COMPLETE")  
    logging.info(f"Compared {original_image_path} with {generated_image_path}")  
    logging.info(f"Current description: {description}")  
    logging.info(f"Updated description: {updated_description}")  

    return updated_description  
 \end{lstlisting}  

\subsection{Zero-shot GPT4V Image Caption Prompt}
\label{sec:zero-shot-gpt4v-prompt}
\begin{lstlisting}[language=]
Describe this image in detail. Don't refer to 'This image' or 'This picture'. Just describe what you see in short, simple terms, but be as specific as possible. Consider the following categories while describing:  
- Feature Correspondence: Distinct features (edges, corners, textures, etc.)  
- Geometric Consistency: Spatial relationships between features  
- Photometric Consistency: Appearance in terms of lighting, color, and intensity  
- Style Consistency: Image styles (photograph, painting, drawing, diagram, medical image)  
- Semantic Consistency: Objects and their parts maintaining their identity and meaning  
- Structural Integrity: Overall structure of the objects in the images

Note:
- IMPORTANT: The above tips are useful for natural images. For graphical, statistical, or diagrammatic images, focus on the data itself and what kind of reasoning is being conveyed.
- Keep overall response to about 130 words or less. Feel free to shorter phrases or incomplete sentences, if it helps to include important details.
\end{lstlisting}

\subsection{Sample Cycle Descriptions}
\label{appendix-sample-cycle-descriptions}

Application of CyclePrompt for image captioning yields detailed descriptions. In each cycle, the discriminator provides feedback about differences between the original image, and the current cycle's image generation. Based on those differences, the system refines the caption to become more and more descriptive. Below are descriptions generated for the examples shown in Figs.~\ref{fig:io-vqav2} and \ref{fig:io-figureqa}.

\underline{Figure~\ref{fig:io-vqav2} VQAv2 Cycle}
\begin{lstlisting}[language=]
{"cycle": 0, "text": "There are several green apples in a white bowl. The apples have small brown spots and the one in the front has a small indentation near the stem. The background is dark and out of focus."}
{"cycle": 1, "text": "Several bright green apples with a smooth, shiny texture are placed in a white bowl with a wide rim. The apples are unblemished, except for one in the foreground that has a small, dark indentation near the stem. The bowl sits on a dark surface, and the background is a blurred, dark brown, providing a stark contrast to the vibrant green of the apples."}
{"cycle": 2, "text": "Several bright green apples with a smooth, shiny texture are placed in a white bowl with a wide rim. The apples are unblemished, except for one in the foreground that has a small, dark indentation near the stem. The bowl sits on a dark surface, and the background is a blurred, dark brown, providing a stark contrast to the vibrant green of the apples. The apples are closely packed together, with one apple prominently in the foreground, slightly obscuring the apples behind it. The lighting is soft and diffused, highlighting the apples' texture and color."}
{"cycle": 3, "text": "Several bright green apples with a smooth, shiny texture are placed in a white bowl with a wide rim. The apples are unblemished, except for one in the foreground that has a small, dark indentation near the stem. The bowl sits on a dark surface, and the background is a blurred, dark brown, providing a stark contrast to the vibrant green of the apples. The apples are closely packed together, with one apple prominently in the foreground, slightly obscuring the apples behind it. The lighting is soft and diffused, highlighting the apples' texture and color. The apples have visible white speckles, and the bowl has a subtle shadow cast on the right side."}
{"cycle": 4, "text": "Several bright green apples with a smooth, shiny texture are placed in a white bowl with a wide rim. The apples are unblemished, except for one in the foreground that has a small, dark indentation near the stem. The bowl sits on a dark surface, and the background is a blurred, dark brown, providing a stark contrast to the vibrant green of the apples. The apples are closely packed together, with one apple prominently in the foreground, slightly obscuring the apples behind it. The lighting is soft and diffused, highlighting the apples' texture and color. The apples have visible white speckles, and the bowl has a subtle shadow cast on the right side. The apples appear more matte than glossy, and the bowl's rim is thick and slightly curved outward."}
\end{lstlisting}

\underline{Figure~\ref{fig:io-figureqa} FigureQA Cycle}
\begin{lstlisting}[language=]
{"cycle": 0, "text": "A horizontal bar graph with eight bars in different colors, each labeled with a color name on the left side. The bars are arranged from top to bottom in the following order: Dark Cyan, Sky Blue, Deep Sky Blue, Chocolate, Deep Pink, Dim Gray, Medium Periwinkle, and Rebecca Purple. The x-axis is labeled \"xaxis label\" and has a scale from 0 to 100. The y-axis is labeled \"yaxis label.\" The graph has a title at the top that reads \"title.\""}
{"cycle": 1, "text": "A horizontal bar graph with eight bars in different colors, each labeled with a color name on the left side. The bars are arranged from top to bottom in the following order: Dark Cyan, Sky Blue, Deep Sky Blue, Chocolate, Deep Pink, Dim Gray, Medium Periwinkle, and Rebecca Purple. The x-axis is labeled \"xaxis label\" and has a scale from 0 to 100. The y-axis is labeled \"yaxis label.\" The graph has a title at the top that reads \"title.\" The bars have varying lengths representing different values on the x-axis, with Dark Cyan being the longest and Rebecca Purple being the shortest."}
{"cycle": 2, "text": "A horizontal bar graph with eight bars in distinct colors, each labeled with a color name on the left side. The bars are arranged from top to bottom in the order of Dark Cyan, Sky Blue, Deep Sky Blue, Chocolate, Deep Pink, Dim Gray, Medium Periwinkle, and Rebecca Purple. The x-axis is labeled \"xaxis label\" with a scale from 0 to 100, and the y-axis is labeled \"yaxis label.\" The graph has a title at the top that reads \"title.\" The bars have varying lengths representing different values on the x-axis, with Dark Cyan being the longest and Rebecca Purple being the shortest. The graph is a clear, 2D representation with no grid lines, and the bars are solid with no patterns or textures."}
{"cycle": 3, "text": "A horizontal bar graph with eight bars in distinct colors, each labeled with a color name on the left side. The bars are arranged from top to bottom in the order of Dark Cyan, Sky Blue, Deep Sky Blue, Chocolate, Deep Pink, Dim Gray, Medium Periwinkle, and Rebecca Purple. The x-axis is labeled \"xaxis label\" with a scale from 0 to 100, and the y-axis is labeled \"yaxis label.\" The graph has a title at the top that reads \"title.\" The bars have varying lengths representing different values on the x-axis, with Dark Cyan being the longest and Rebecca Purple being the shortest. The graph is a clear, 2D representation with no grid lines, and the bars are solid with no patterns or textures. The title, axis labels, and color labels are all clearly legible."}
{"cycle": 4, "text": "A horizontal bar graph with eight bars in distinct colors, each labeled with a color name on the left side. The bars are arranged from top to bottom in the order of Dark Cyan, Sky Blue, Deep Sky Blue, Chocolate, Deep Pink, Dim Gray, Medium Periwinkle, and Rebecca Purple. The x-axis is labeled \"xaxis label\" with a scale from 0 to 100, and the y-axis is labeled \"yaxis label.\" The graph has a title at the top that reads \"title.\" The bars have varying lengths representing different values on the x-axis, with Dark Cyan being the longest and Rebecca Purple being the shortest. The graph is a clear, 2D representation with no grid lines, and the bars are solid with no patterns or textures. The title, axis labels, and color labels are all clearly legible. The graph background is white, and the bars are not stacked."}
\end{lstlisting}

\subsection{Relative Complexity: Continued}

\begin{figure} 
  \centering 
  \includegraphics[width=\linewidth]{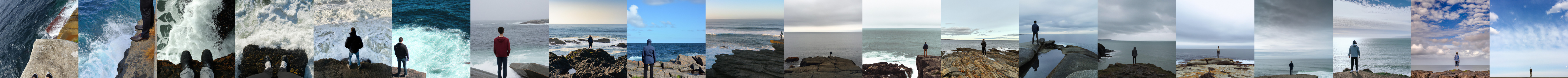}  
  \caption{Text-Image-Text cycle for ``Standing at the edge"}  
  \label{fig:text-image-text-standing-at-the-edge}  
\end{figure}

\begin{figure}  
\centering  
  \includegraphics[width=\linewidth]{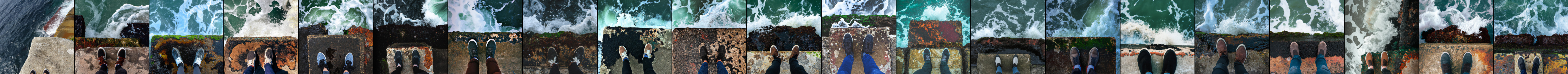}  
  \caption{Image-Text-Image cycle on first generation of Fig. \ref{fig:text-image-text-standing-at-the-edge}}  
  \label{fig:image-text-image-standing-at-the-edge}  
\end{figure}

We provide additional examples of the Text-Image-Text (image generation) and Image-Text-Image (image captioning) cycles.When the input space is lower-complexity (e.g. text to image, in Fig.\ref{fig:text-image-text-standing-at-the-edge}), the output space can comply with the input while continuously changing. When the input space is higher-complexity (e.g. image to text, in Fig. \ref{fig:image-text-image-standing-at-the-edge}), both spaces are constrained and converge.

\end{document}